# Super Resolution Convolutional Neural Network Models for Enhancing Resolution of Rock Micro-CT Images

Wang YD[1], Armstrong R. T.[1], Mostaghimi P[1]

1 School of Minerals and Energy Resources Engineering, The University of New South Wales, NSW 2052, Australia

Highlights:

- Super Resolution CNN on µCT images in the DRSRD1 Dataset of digital rocks
- SR-Resnet, EDSR, and WDSR models show +3-5 dB in quality vs bicubic interpolation
- Edge sharpness is completely recovered with high frequency noise related loss
- SRCNN can both upscale and precondition for image segmentation and other processes

## 1 ABSTRACT

In this paper, we apply and compare Single Image Super Resolution (SISR) techniques based on Super Resolution Convolutional Neural Networks (SRCNN) on micro-computed tomography (µCT) images of sandstone and carbonate rocks. Digital rock imaging is limited by the capability of the scanning device resulting in trade-offs between resolution and field of view, and super resolution methods tested in this study aim to compensate for these limits. SRCNN models SR-Resnet, Enhanced Deep SR (EDSR), and Wide-Activation Deep SR (WDSR) are used on the Digital Rock Super Resolution 1 (DRSRD1) Dataset of 4x downsampled images, comprising of 2000 high resolution (800x800) raw micro-CT images of Bentheimer sandstone and Estaillades carbonate. The trained models are applied to the validation and test data within the dataset and the Peak Signal to Noise Ratio compared against typical interpolation methods, showing a 3-5 dB rise in image quality compared to bicubic interpolation, with all tested models performing within a 0.1 dB range. WDSR is applied to Bentheimer, Berea, and Leopard sandstones, and Savonnieres carbonate. Difference maps indicate that edge sharpness is completely recovered in images within the scope of the trained model, with only high frequency noise related detail loss. We find that aside from generation of high-resolution images, a beneficial side effect of super resolution methods applied to synthetically downgraded images is the removal of image noise while recovering edgewise sharpness which is beneficial for the segmentation process. The model is also tested against real low-resolution images of Bentheimer rock with image augmentation to account for natural noise and blur. The SRCNN method is shown to act as a preconditioner for image segmentation under these circumstances which naturally leads to further future development and training of models that segment an image directly. Image restoration by SRCNN on the rock images is of significantly higher quality than traditional methods and suggests SRCNN methods are a viable processing step in a digital rock workflow.

# 2 INTRODUCTION

X-ray micro-computed tomography (micro-CT) is an imaging technique that allows for the generation of three-dimensional images that detail the micro-structure of porous rock. These images can be used for determination of petrophysical and flow properties of rocks (Mostaghimi, Blunt et al. 2013, Krakowska, Dohnalik et al. 2016, Chung, Wang et al. 2018, Wang, Chung et al. 2019). It is a non-invasive and non-destructive method which allows rock samples to be imaged and subsequently used in laboratory if needed (Hazlett 1995, Lindquist, Lee et al. 1996, Wildenschild and Sheppard 2013, Schlüter, Sheppard et al. 2014). X-ray micro-CT scanners can now image rocks sample down to a few micrometres, which is sufficient to allow the micro-structure of conventional rocks to be characterised (Flannery, Deckman et al. 1987, Coenen, Tchouparova et al. 2004). In order to obtain fine detail and a more representative image of a rock sample for flow simulation or otherwise, a high-resolution micro-CT image of a rock sample is required. However, these high resolution images are typically of small field of view leading to non-representative results (Li, Teng et al. 2017). To have an image with high resolution and large field of view, it is possible to apply super resolution algorithms on low resolution data. Aside from traditional interpolation methods (nearest neighbour, linear, bicubic, etc), the sparse coding example based SR method has shown superior results but is inflexible and computationally intensive during the generation phase (Yang, Wright et al. 2008).

Convolutional neural networks (CNNs) have successfully been used as a learning-based algorithm for generating Single Image Super Resolution (SISR) (Dong, Loy et al. 2014, Wang, Wang et al. 2016, Ledig, Theis et al. 2017, Lim, Son et al. 2017, Yu, Fan et al. 2018) under the umbrella term Super Resolution Convolutional Neural Networks (SRCNNs). Typically, the algorithm is applied to the problem of generating a super resolution (SR) image from a low resolution (LR) image that is a bicubic downsample from an unseen high resolution (HR) image. In the context of micro-CT images, the application of SRCNN methods based on (Dong, Loy et al. 2014) have been implemented for both medical-CT, medical X-rays, and micro-CT images of rocks as part of a Digital Rock Physics (DRP) workflow (Umehara, Ota et al. 2018, Wang, Teng et al. 2018), while more advanced SRCNN models have yet to be applied to DRP.

Super resolution recovery from a low resolution image is an undetermined inverse problem, with an infinite number of possible outcomes (Dong, Loy et al. 2014). CNN hidden layers are used as an implicit version of the sparse coding based model (Li, Teng et al. 2017), and provide better flexibility with diverse or specific example datasets. SRCNN models all follow a similar architecture, with 2 distinct formulations whereby the input image is either the original LR image (Wang, Wang et al. 2016, Ledig, Theis et al. 2017, Lim, Son et al. 2017, Yu, Fan et al. 2018), or a pre-processed upscaled image (usually by bicubic interpolation) (Dong, Loy et al. 2014, Umehara, Ota et al. 2018, Wang, Teng et al. 2018). The pre-processed models were first developed (Dong, Loy et al. 2014) and were designed as shallow CNNs with a depth of 3-5 layers. Later implementations increased the model depth (Kim, Kwon Lee et al. 2015) , with the use of skip connections to preserve feature maps from shallower layers, with depth ranging from 8-32 layers.

The simplest and first SRCNN models, aptly named SRCNN (Dong, Loy et al. 2014), applied bicubic upsampling (BC) as a pre-processing step, with convolutional layers producing a BC to HR mapping. This has been superseded by more integrated LR to HR methods due to the inefficient need for a fully sized input image that increases the computational cost by the scaling factor multiplied by the dimensionality of the image. Recent methods have shifted from using deconvolutional layers at the end of the network which caused checkerboard artifacting, to the current preferred method of pixel-shuffling the convolutional filters (subpixel convolution) (Yu, Fan et al. 2018). Deep learning methods, as suggested by its title, performs better the deeper the model and SRCNN methods utilise model depth while preserving important shallow feature sets by utilising the skip connection, that adds outputs from shallow layers to deeper layers (Ledig, Theis et al. 2017). Batch normalisation, while a common feature in many deep networks, and present in the SR-Resnet model (Ledig, Theis et al. 2017), have been empirically observed to reduce the accuracy of SRCNN methods that rely on mini batches of cropped images (due to the highly variant batch characteristics), and as such, are not present in more recent formulations (Lim, Son et al. 2017). SRCNN methods can only regenerate subpixel features that were present in the original HR images. In the context of DRP, microporosity can only be regenerated from LR images if the SRCNN model was trained on HR images that had resolved micropores. Furthermore, since image segmentation is a major aspect of DRP workflows, SRCNN models that possess some form of intrinsic noise suppression while maximising edge recovery are favourable outcomes.

Super Resolution Generative Adversarial Network (SRGAN) methods that combine traditional SRCNN with image classification CNN have been shown to recover high resolution features in a perceptual manner. The SRGAN results in features that look to the human eye as realistic when surveyed, but are in fact lower in pixel by pixel accuracy (Ledig, Theis et al. 2017). This is not necessarily beneficial to super resolution recovery of micro-CT images, as the generation of pixelwise fake high frequency data is non-deterministic. In noisy micro-CT images, the regenerated high frequency data could either be unwanted image noise or useful sub-pixel features with no obvious way to distinguish them. This problem is less prominent in the use of GANs in photorealistic SR applications, since the LR to HR mapping is performed on less noisy digital photographs that result in generated high frequency information more closely resembles real features.

In this paper, we train SRCNN models developed with an "end-to-end" architecture (Wang, Wang et al. 2016), shown to be superior to the original SRCNN method, including the SR-Resnet, EDSR, and WDSR models with the DRSRD1 dataset. These models represent different implementations of SRCNN, with SR-Resnet representing a traditional Deep CNN, EDSR an optimised simplification of the SR-Resnet model, and WDSR a wider but shallower CNN implementation. The image similarity metric Peak Signal to Noise Ratio (PSNR) for all models and their variants is compared. We find that there is a minimal difference in the achieved PSNR per model, but note that WDSR models tend to perform slightly better, while requiring only a fraction of the required model parameters. The best performing model (WDSR-B w/ 32 layers) is then used to super resolve unseen digital rock images. These include Bentheimer, Berea, and Leopard sandstones, and Savonnieres carbonate to encompass a typical range of common digital rocks. The model is tested with noise augmentation on low resolution Bentheimer images with natural noise to investigate the impact of image noise on SR, and finally tested on the original HR data to observe the image characteristics of HR-SR images and the flexibility of the model.

# 3   MATERIALS AND METHODS

## 3.1   DATASETS

The DRSRD1_ 2D dataset (Wang, Armstrong et al. 2019) is used in this study. It comprises of 2000 800x800 high resolution unsegmented slices of Bentheimer sandstone [3.8 micrometres] and Estaillades carbonate [3.1 micrometres] images. The images are synthetically downsampled by a factor of 4x using the matlab imresize function, commonly used for SR datasets (Agustsson and

Timofte 2017), with one set using only the "bicubic" downsample method, while the "unknown" dataset applies a random sampling of methods including box, triangle, cubic, lanczos2, and lanczos3.

Models trained on the DRSRD1 dataset are validated against rock images that are external to the dataset, including Bentheimer, Berea, and Leopard Sandstones, and Savonnieres carbonate. A lower resolution micro-CT image of Bentheimer [7 microns] with natural high frequency noise is used to illustrate the noise suppression characteristics of SRCNN methods.

*Table 1: Datasets used for SRCNN testing*

|  | Source | Resolution (microns) | Reference |
|---|---|---|---|
| DRSRD1 Sandstone | DRSRD1 | 3.8 | (Wang, Armstrong et al. 2019) |
| DRSRD1 Carbonate | Estaillades Carbonate #2 | 3.1 | (Bultreys 2016, Wang, Armstrong et al. 2019) |
| External Bentheimer | Bentheimer: 98% air saturation | 4.9 | (Herring, Sheppard et al. 2018) |
| External Berea | Berea: 94% air saturation | 4.6 | |
| External Leopard | Leopard: dry | 3.5 | |
| External Savonnieres Carbonate | Micro-CT scan Savonnieres | 3.8 | (Bultreys 2016) |
| Low Resolution Bentheimer | Bentheimer | 7 | (Ramstad 2018) |

All training was performed on a GTX1080ti Nvidia GPU using the TensorFlow library.

All models were trained on cropped images (192 x 192) from the DRSRD1 shuffled2D dataset. Since the training is performed on randomly sampled cropped images (48x48 low resolution and 192x192 high resolution) in batches of 16 such images, the model PSNR is validated on the DRSRD1 validation dataset every 1000 iterations (forming a training epoch).

## 3.2 SUPER-RESOLUTION CNN

The architecture in this study takes a single 2D LR image $X$ and will generate an SR image $F[X]$ such that it is as similar as possible (on a pixel-by-pixel basis) as the original HR image $Y$. In this case for a scaling factor of 4 (commonly the upper tested limit for SRCNN methods), the image $X$ measures $\frac{Nx}{4}$ by $\frac{Ny}{4}$ where $Nx$ and $Ny$ are the dimensions of the original high-resolution image. In general, this particular type of architecture will attempt to generate an SR image from a LR image at fixed scaling factors.

In general, the methods employed in this study are based on a formulation that applies upsampling and pixel-binning on the convolutional filters in the deepest layers of the model. The basic architecture outlined below illustrates the comparative structure, with upsampling layers occurring in the deeper layers, utilising the filter information as opposed to applying a fixed bicubic interpolation at the start of the model which causes the model to bloat up in computational cost.

The CNN architectures used can be summarised with the following steps:

1. Input LR images
2. Convolution and Activation

a) An input array $X$ is convolved against a set of convolutional filters $W$ which has a shape of $n_f \times k_x \times k_y$ where $n_f$ is the number of filters and k is the size of each filter (kernel size) and shifted by biases $b_f$. The output $F[X]$ is passed through a Rectified Linear Unit (ReLU) activation function that takes the form $\max(0, Y) - \alpha \max(0, -Y)$ where $\alpha$ is equal to 0 (ReLU), a constant (Leaky ReLU [LReLU]), or a learnable variable (Parametric ReLU [PReLU]) (He, Zhang et al. 2015).
3. Residual Layers
    a) These are repeated sets of layers in the architecture that form the "depth" of the model. A typical residual block contains activated convolutional layers of various configurations.
4. Convolution [optional]
5. Skip Connections
    a) A skip connection adds together the output from the shallow layers $Y_{shallow}$ with the output from the deep layers after the residual blocks $Y_{deep}$ to improve gradient scaling at deeper layers which tend to suffer from numerically vanishing gradients that are difficult to iterate.
6. Convolution [optional]
7. Upsampling/Deconvolve/Subpixel convolution
    a) The output from the shallow and deep layers once added together are upsampled to the same width and height as the HR images. The upsampling algorithm is typically a subpixel convolution (depth to space transform) that converts an array of size $[X, Y, Z]$ into size $\left[nX, nY, \frac{Z}{n^2}\right]$ where $n$ is the scaling factor.
8. Output SR images

## SR-Resnet

The SR-Resnet architecture (Figure 1) used in this study applies a wide initial convolution layer of kernel size 9 with PReLU activation. All convolutional layers apply 64 filters. This stage of the network is saved for use in a later skip connection. A series of 16 residual blocks are added onto each other with each block containing in order; a convolutional filter of kernel size 3, batch normalisation, PReLU, another convolutional layer, and another batch normalisation. This output is added to the input of the residual block. After the residual blocks, another convolutional layer is applied with batch normalisation. The skip connection is applied, adding the residual output to the initial layer. The subpixel convolutions are applied, with a convolutional layer of kernel size 3 and 256 filters followed by sub pixel convolution with PReLU. This is applied twice for a scale factor of 4. A final convolutional layer with 1 or 3 filters (depending on if images are greyscale or RGB) is applied to recover the SR image.

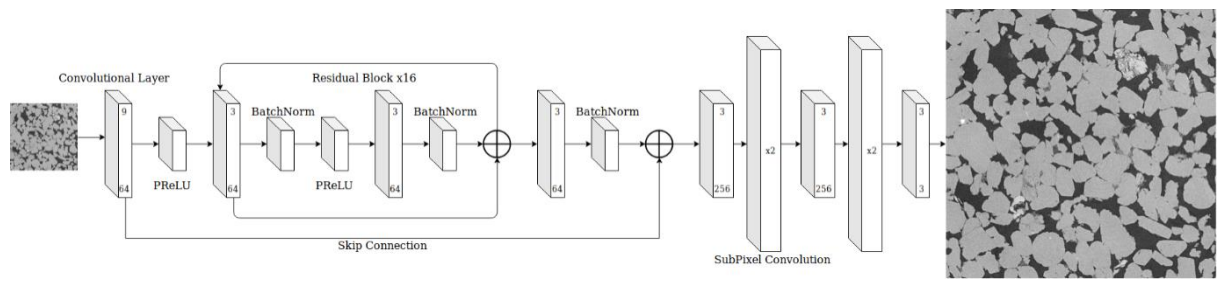

Figure 1: Architecture of the SR-Resnet SRCNN model

## Enhanced Deep Super Resolution (EDSR)

The EDSR architecture (Figure 2) is similar the SR-Resnet model, but with some simplifications. The input convolutional layer is not wide (with a typical kernel size of 3), all batch normalisation layers are removed, and all PReLU activation layers are removed. A reLU activation is applied between the residual block convolutional layers. In this study, EDSR models with 16 and 32 residual blocks are tested on the DRSRD1 dataset

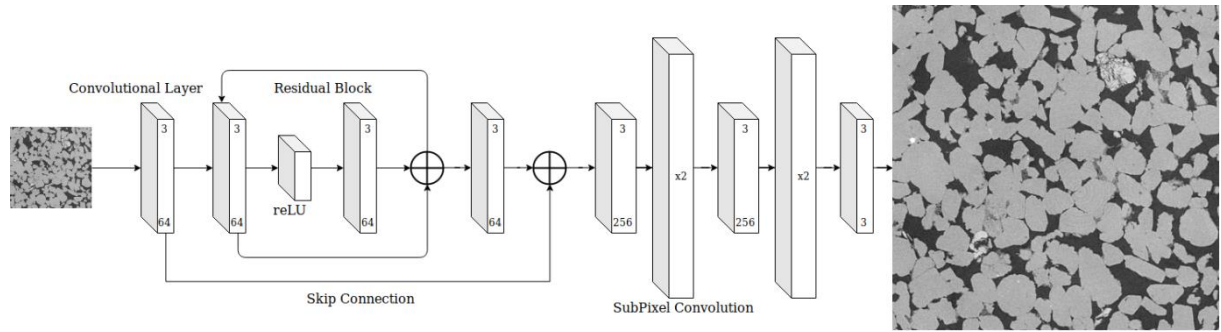

*Figure 2: Architecture of the EDSR SRCNN model*

Wide Activation Deep Super Resolution (WDSR)

In the WDSR models the number of convolutional filters is reduced to 32 and the skip connection and residual blocks are distinctly different to the SR-Resnet and EDSR models. In the WDSR-A model (Figure 3), residual blocks are structurally identical to the EDSR residual block with the only difference being that the number of filters in the first convolutional layer in block is expanded by a factor of 4 (totalling 128).

The residual block in WDSR-B (Figure 4) contains a convolutional layer with a filter expansion factor of 6 (totalling 192) with reLU activation and a kernel size of 1. This is followed by another convolutional layer of kernel size 1, with 154 layers (80% of 192). The block ends with a convolutional layer with kernel size 3 and residual block addition.

The pixel shuffle routine in the WDSR model used in this study is coupled with the skip connection by applying subpixel convolution to both the residual block output and the initial layer before adding them together. Before the subpixel convolution, the residual block outputs are passed through a convolutional layer with 48 filters and kernel size of 3. The initial layer is passed through a convolutional layer with 48 filters and a kernel size of 5.

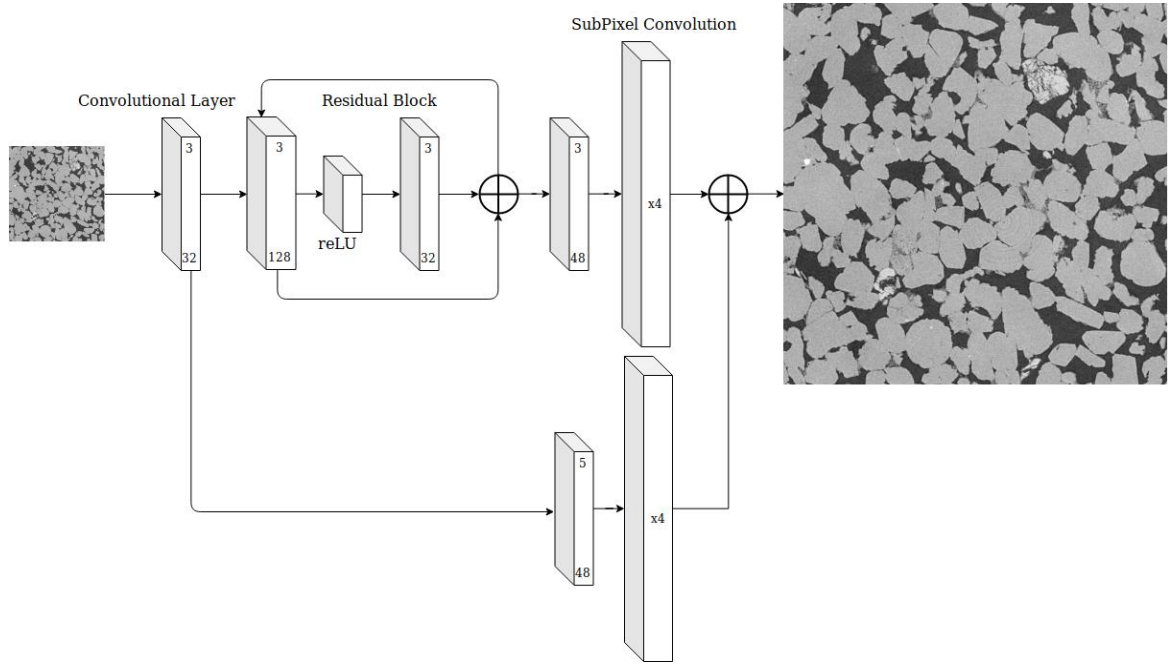

*Figure 3: Architecture of the WDSR-A SRCNN model*

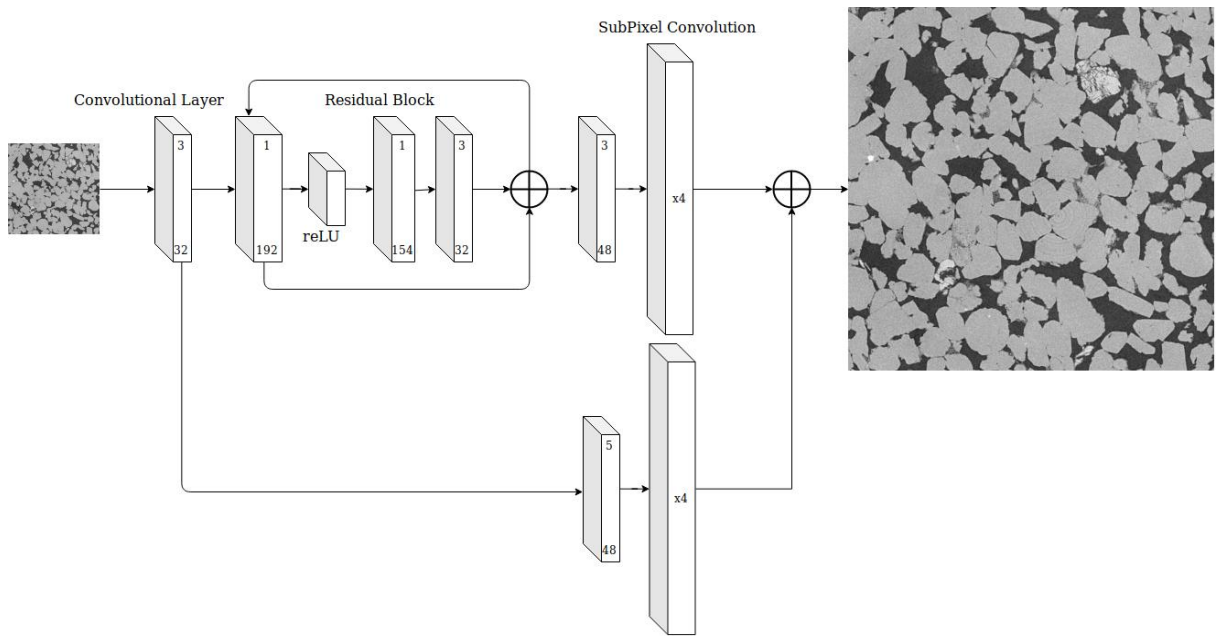

*Figure 4: Architecture of the WDSR-B SRCNN model*

Loss and Metrics

Using the Adam optimiser (Kingma and Ba 2014) on the MSE (Eqn ( 3 )), the learning rate (which dictates the speed/stability trade-off of the optimiser) for SRCNN has been shown to be stable around 1e-4 with reduced stability at higher iteration counts. With weight normalisation (Salimans and P. Kingma 2016), the WDSR learning rate is stable at 1e-3. A decaying learning rate $lr$ is applied with a half-life given by Eqn ( 1 ):

$$lr_i = lr_{init}(0.5)^{\frac{epoch}{step}} \qquad (1)$$

The Peak Signal-to-Noise Ratio (PSNR) is calculated by Eqn ( 2 ):

$$PSNR = 10\log_{10}\frac{[\max(I_1, I_2) - \min(I_1, I_2)]^2}{MSE} \qquad (2)$$

where $I_1$ and $I_2$ are the pixels within the images to compare, and $MSE$ is the mean squared error, calculated by Eqn ( 3 ):

$$MSE = \frac{1}{N}\sum_{i=1}^{N}(I_{1_i}^2 - I_{2_i}^2) \qquad (3)$$

where $N$ is the number of pixels in the images. The PSNR metric is measured in dB and a convenient image comparison parameter. The MSE calculates the pixelwise difference between 2 images and is the loss function used in SRCNN. The SRCNN essentially minimises the MSE between SR and HR images by tuning the model parameters such as the convolutional filter values using the Adam algorithm. This is analogous to other matching algorithms such as linear regression, history matching, and implicit finite difference.

## 4 RESULTS AND DISCUSSION

The SR-Resnet, EDSR, and WDSR models are generated and trained with varying parameters, as listed in Table 2. The results of the training, reported in PSNR, are shown in Figure 5 indicating similar training performance for all models on both default (bicubic) and unknown LR datasets. The training time and model parameters are also provided, showing a distinct advantage for the WDSR models due to more efficient architecture. SR images of unseen digital rock images are generated to test the model flexibility, including Bentheimer, Berea, and Leopard sandstones, and Savonnieres carbonate, encompassing a typical range of common digital rocks. Noise augmentation is tested on low resolution Bentheimer images with natural noise to investigate the impact of image noise on SR and also tested on the original HR data to observe the image characteristics of HR-SR images.

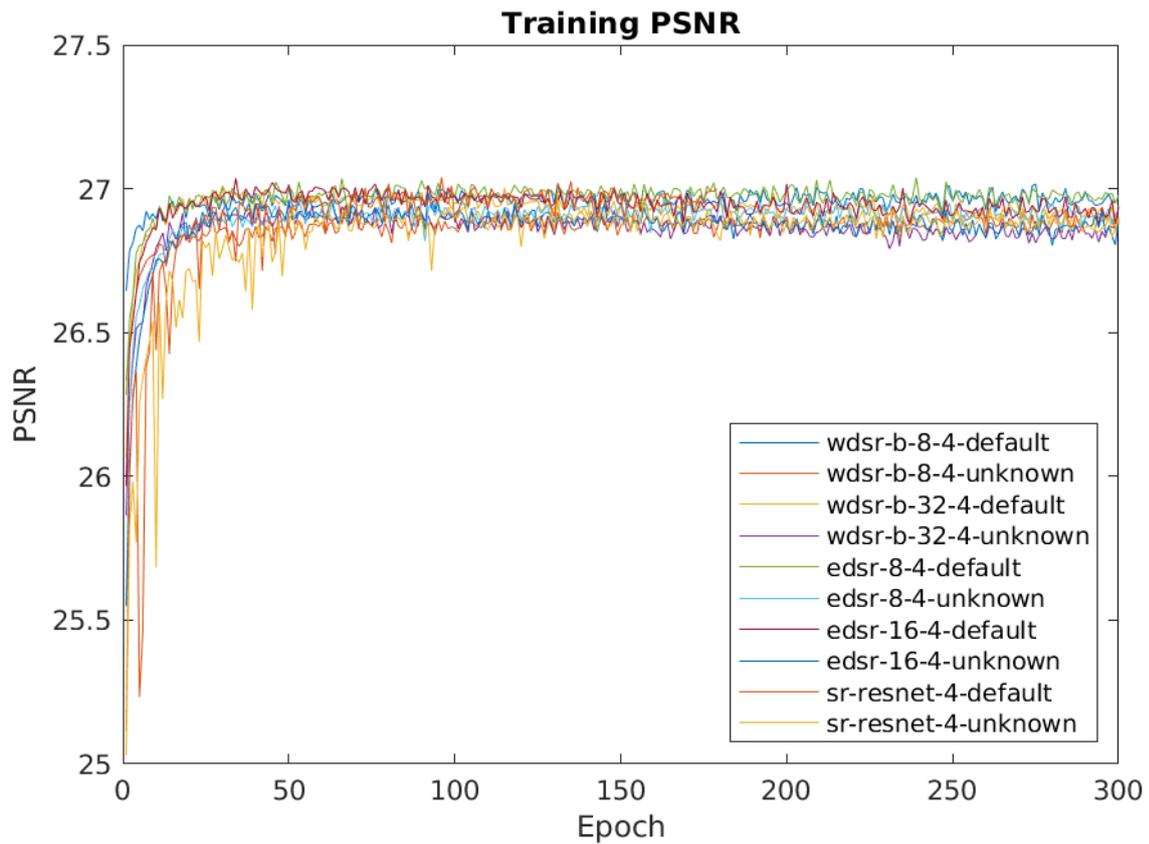

*Figure 5: Plot of training results for the tested models based on pixelwise accuracy. All models reach plateau around 50-100 epochs. Each model and its parameters are outlined in Table 2*

### 4.1 IMAGE SIMILARITY METRICS

Of the 300 epochs trained for each model variant, the validation takes the epoch which gives the highest PSNR when calculated against the shuffled2D validation folders which comprise of 200 unseen, full-size images of the Bentheimer sandstone and Estaillades carbonate (100 each at 800x800). The results of the validation for each epoch is graphed in Figure 6 and the best achieved models are also tabulated in Table 2.

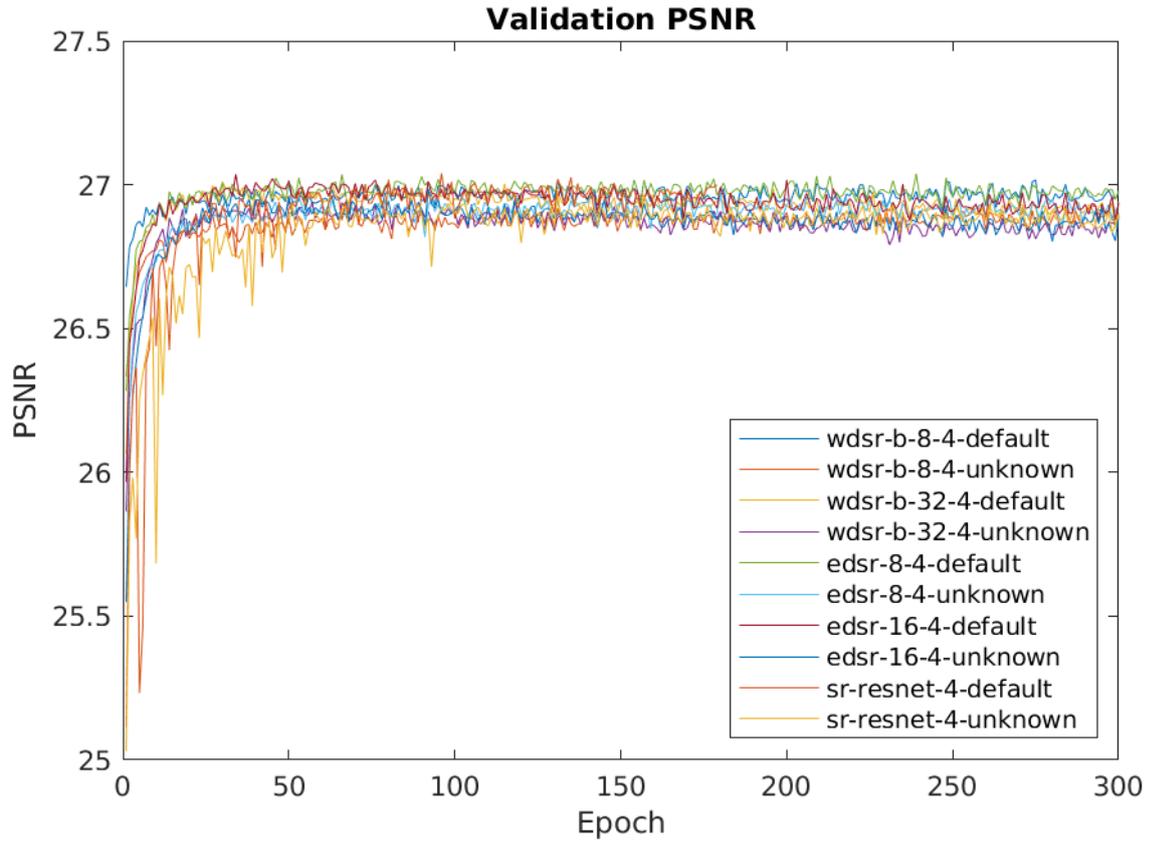

*Figure 6: PSNR achieved by each model when applied to the full-size validation set. The validation PSNR values are similar to the training results despite training being performed on cropped images (192x192) and validation being performed on full images (800x800). This is likely due to the main features of micro-CT rock images (pore, solid, edge, noise) being present in the cropped images.*

*Table 2: Performance data for different SRCNN models applied to the bicubic and unknown datasets from DRSRD1. Improvements in model performance between SR-Resnet and EDSR (mostly due to the removal of batch normalisation) are apparent. The WDSR models outperform based on an architecture that captures wider features. The best models obtained for the unknown and bicubic datasets are highlighted red.*

| Model | No. Residual Layers | Training Time [hrs] | Data Subset | Best Epoch | Validation PSNR | Bicubic PSNR |
|---|---|---|---|---|---|---|
| EDSR | 8 | 5 | bicubic | 68 | 26.9770 | 24.0716 |
| EDSR | 8 | 5 | unknown | 101 | 26.9147 | 23.9472 |
| EDSR | 16 | 7 | bicubic | 37 | 26.9744 | 24.0716 |
| EDSR | 16 | 7 | unknown | 55 | 26.9116 | 23.9472 |
| SR RESNET | 16 | 8.5 | bicubic | 58 | 26.9701 | 24.0716 |
| SR RESNET | 16 | 8.5 | unknown | 89 | 26.9014 | 23.9472 |
| WDSR A | 8 | 3 | bicubic | 28 | 26.9747 | 24.0716 |
| WDSR A | 8 | 3 | unknown | 29 | 26.9070 | 23.9472 |
| WDSR A | 32 | 9 | bicubic | 29 | 26.9627 | 24.0716 |
| WDSR A | 32 | 9 | unknown | 31 | 26.9049 | 23.9472 |
| WDSR B | 8 | 3.5 | bicubic | 103 | 26.9881 | 24.0716 |
| WDSR B | 8 | 3.5 | unknown | 203 | 26.9093 | 23.9472 |
| WDSR B | 32 | 10 | bicubic | 41 | 26.9875 | 24.0716 |
| WDSR B | 32 | 10 | unknown | 44 | 26.9307 | 23.9472 |

Validation of the trained models shows that the best performing model on the unknown dataset was the WDSR-B model with 32 base layers, while the best model for the bicubic dataset was the WDSR-B with 8 base layers. However, the differences in overall validation performance are minor, ranging from 26.9070 to 26.9307 for the models trained on the unknown dataset. All of the models implemented and tested in this study use subpixel convolution to achieve LR to SR mappings which when compared to older methods such as deconvolution and BC pre-processing, gives the largest performance gains. The removal of batch normalisation from SR-Resnet to EDSR gives a minor PSNR boost from 26.9014 to 26.9116. this is consistent across the EDSR results, with all 4 tested models outperforming the SR-Resnet model for the respective trained datasets. The WDSR-A and WDSR-B models show mixed results against EDSR, with some outperforming significantly while others are slightly worse. Nevertheless, the reduced number of parameters in the WDSR models gives a significant performance boost, with the 8-layer WDSR-B and EDSR requiring 3.5 and 5 hours respectively for 300,000 iterations.

To apply SRCNN in a general manner, the training results obtained from the unknown datasets are used for further evaluation. Real low resolution micro-CT images are unlikely possess the same LR characteristics as those obtained from bicubic downsampling, shown in the later section 4.2. The best epoch from the WDSR-B-32 model obtained from training on the unknown dataset is used to generate SR images of the combined validation and test datasets for sandstone and carbonate (800 x 800 x 200). The unknown LR datasets are used for further tests as they are more generalised in their LR to HR mapping. The PSNR results for different models tested specifically on the sandstone and carbonate subsets are shown in Table 3.

*Table 3: PSNR model metrics for sandstone and carbonate subsets in the DRSRD1 dataset. The better resolved sandstone images have better SR performance, while the carbonates that are noisy with sub resolution micro-porosity features, are more difficult to super resolve.*

|  | Sandstone |  | Carbonate |  |
| --- | --- | --- | --- | --- |
|  | Mean PSNR | Var PSNR | Mean PSNR | Var PSNR |
| Bicubic | 25.3981 | 0.2166 | 22.5182 | 0.1781 |
| WDSR-B-8 | 29.4199 | 0.0647 | 23.4950 | 0.1635 |
| WDSR-B-32 | 29.4444 | 0.0641 | 23.5152 | 0.1629 |
| EDSR-8 | 29.4343 | 0.0635 | 23.5136 | 0.1621 |
| EDSR-16 | 29.4286 | 0.0645 | 23.5172 | 0.1635 |
| SR-RESNET | 29.4245 | 0.0638 | 23.5352 | 0.1619 |
| WDSR-A-8 | 29.4099 | 0.0650 | 23.5085 | 0.1624 |
| WDSR-A-32 | 29.4263 | 0.0635 | 23.5184 | 0.1618 |

The sandstone images show a comparison between the PSNR obtained from bicubic interpolation of the LR images (PSNR averages 25.3981) and SR images (PSNR averages 29.4444). While there is a minimal difference in the performance of SRCNN methods in terms of averages and variances, they outperform traditional bicubic interpolation. The carbonate images show similar image metrics and features with lower overall PSNR values due to the less distinguishable features of carbonate micro-CT images, which can be misconstrued as image noise. Overall, a PSNR boost from 22.5 to 23.5 compared to bicubic interpolation.

A visual inspection of the testing results shown by the zoomed in images in Figure 7 shows a clear improvement in SR image quality between the bicubic method and the SRCNN methods. Between the SRCNN methods there are visually minimal differences. The loss of high frequency features is evident in all SR methods as they attempt to minimise the pixelwise information loss. The intra-granular noise is blurred out and the bicubic method also attenuates the sharpness of grain edges. The SRCNN methods appear to be able to recover edges very well, which is shown in the difference maps in Figure 8. Sample images from the carbonate testing set are also shown in Figure 9 with similar visual image metrics of a reduction in noise without impacting edges, which is supported by the difference maps depicted in Figure 10.

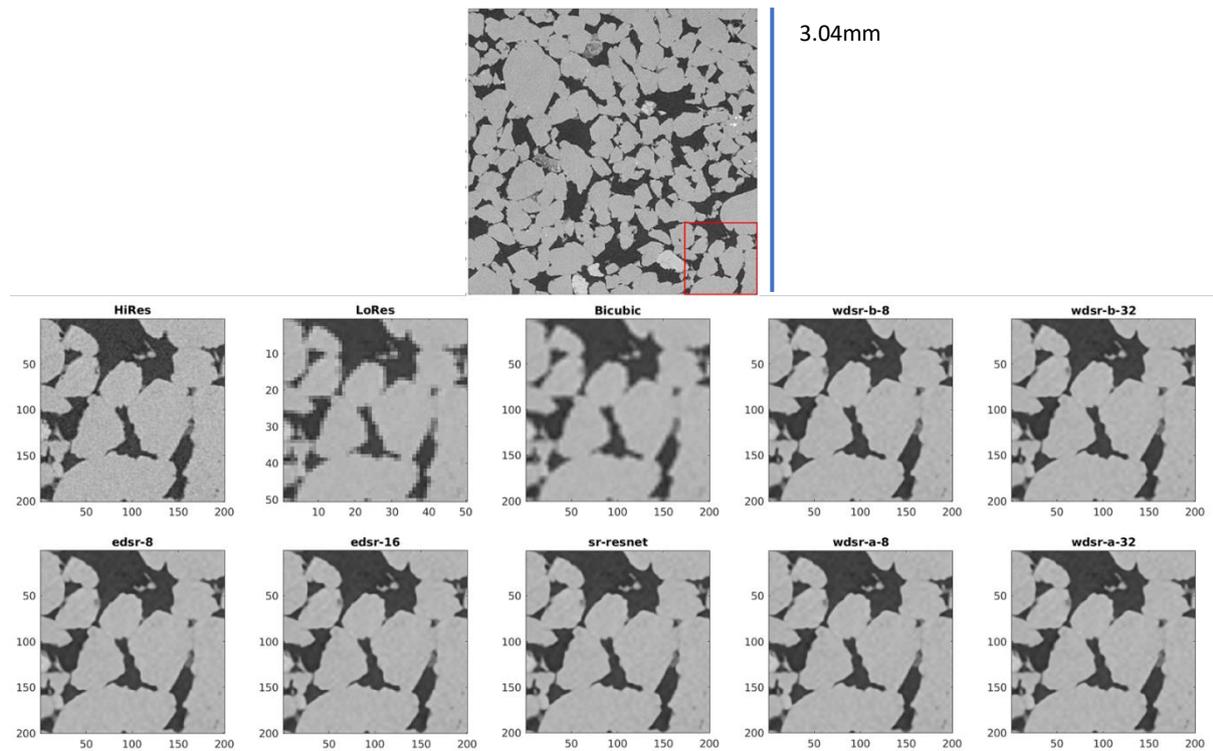

Figure 7: Zoomed in view of the comparison between HR, LR, and SR images. The SRCNN models appear unable to recover high frequency details such as the intra-granular noise but are excellent at reconstructing edge sharpness.

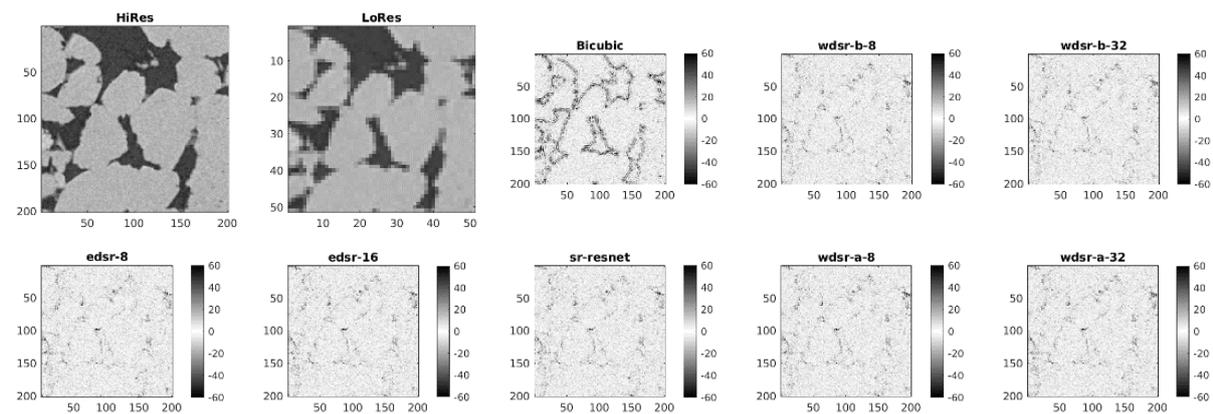

Figure 8: Pixelwise difference maps for the SR sandstone sample image. It can be clearly seen that edges are lost in bicubic interpolation, while SRCNN models tend to recover edges while losing high frequency noise.

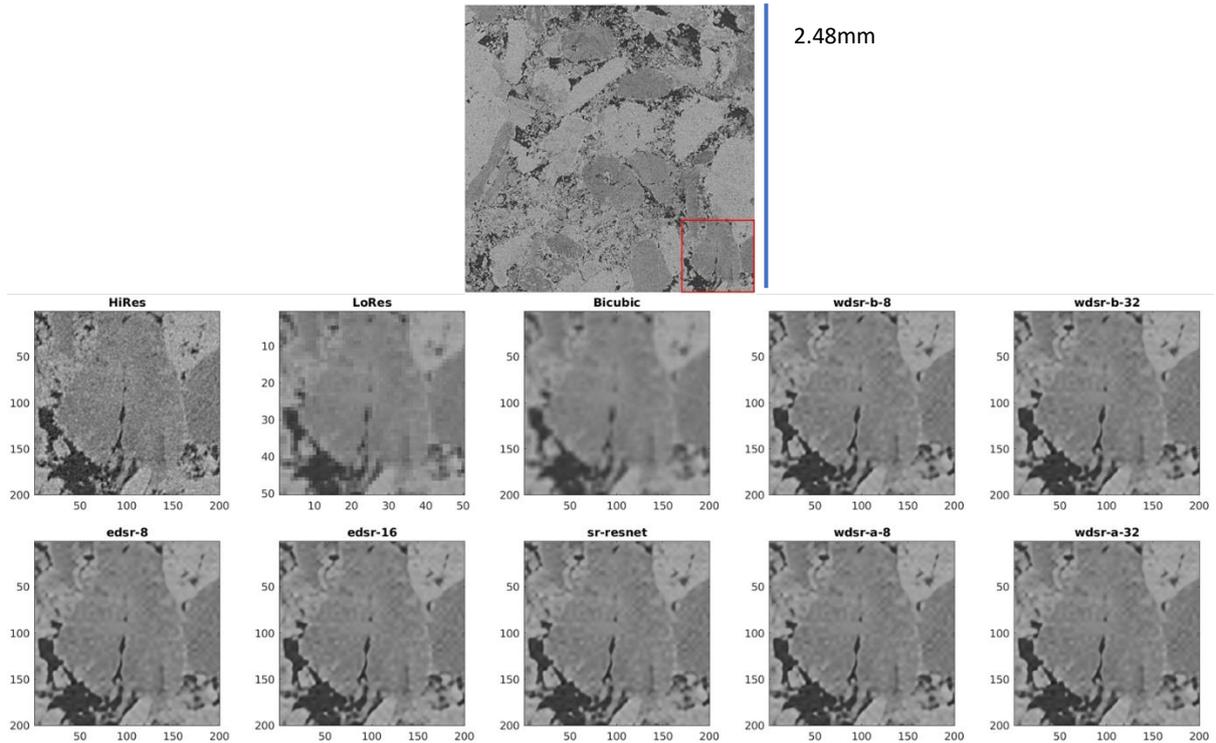

*Figure 9: Zoomed in view of the carbonate SR images. Similar metrics are present in carbonate SR images as those found in sandstone, particularly the loss of intra-granular detail and the recovery of edges.*

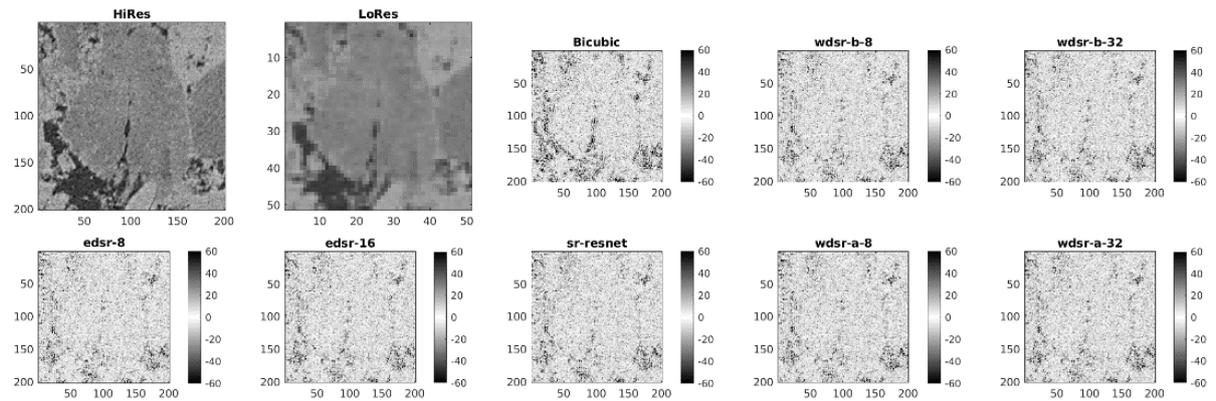

*Figure 10: Carbonate SR difference maps show that edge recovery is not as successful as the sandstone, likely due to the inherently less defined edges in carbonate images.*

Using the trained WDSR-B-32 model on unseen, externally sourced samples of Bentheimer, Berea, Leopard sandstone, and Savonnieres carbonate (see section 3.1), the resulting super resolution PSNR is compared to the bicubic interpolation PSNR. The images are synthetically downsampled by a scale factor of 4 in the same manner as the unknown subset in the DRSRD1 dataset. Figure 11 to Figure 14 show examples of the resulting SR images of Bentheimer, Berea, Leopard sandstones, and Savonnieres carbonate. We choose these images for the following reasons. The DRSRD1 dataset is a Bentheimer rock sample, so another Bentheimer rock image from another CT source is a useful first step in testing model flexibility. Berea and Leopard sandstones represent a further step out of the model training scope but retains typical grain sizes and image characteristics. Savonnieres carbonate is similar to the Estaillades carbonate in that it contains significant sub resolution micro-porosity, but differs in some of its rock morphology, as it is highly oolitic compared to the DRSRD1 carbonate training and testing images.

Visually, all the SR images show noticeable improvement from bicubic interpolation results. However, the overall performance is impacted as the model is untrained on these rock images. Errors in the Berea sample are most prominent, as it contains an aqueous phase of high intensity values that the model is untrained on. Overall, the edge recovery shows good results as most of the difference maps obtained (barring the Berea) show that the pixelwise losses are primarily high frequency noise data that is smaller than the convolutional kernel. The loss of noise features is especially prominent in the Leopard sandstone and the Savonnieres carbonate SR images due to the tighter grain space in the sandstone and the microporosity in the carbonate. The use of pixelwise loss results in a natural outcome of denoising the image while preserving sharpness at edges. While this may be detrimental to the overall "accuracy" of the SR model (hence the use of perceptual losses and SRGANs in photographic SR applications), in the context of micro-CT imaging workflows that later require image segmentation, this is an inadvertent benefit.

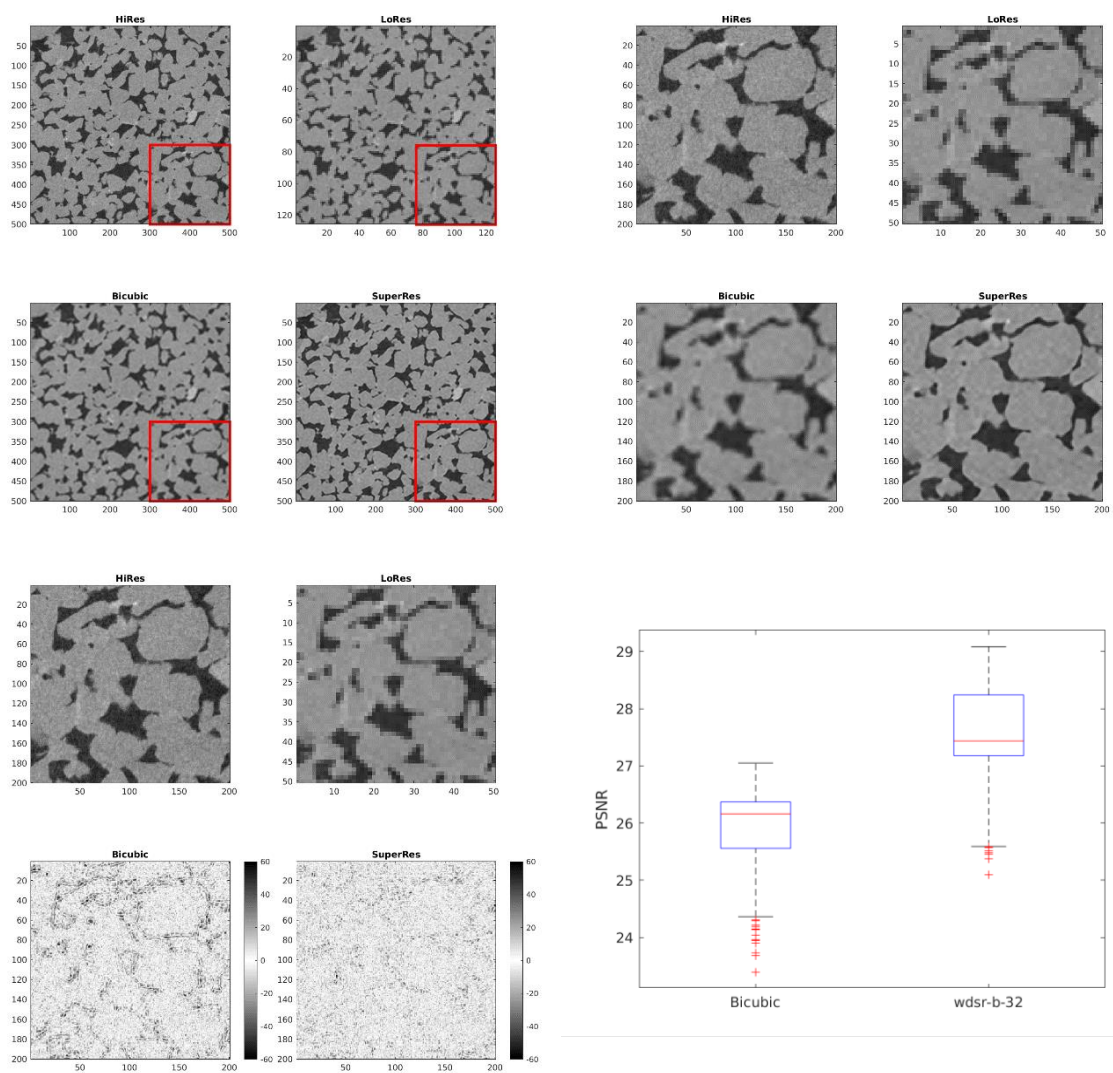

*Figure 11: SR metrics and visualisations of externally sourced Bentheimer Sandstone. The overall performance of the model is reduced, reaching a SR PSNR range around 27-28, as opposed to 28.5-30 on the DRSRD1 dataset, as can be seen from the box plots on the bottom right. Despite this, images on the top and difference maps shown on the bottom left show that the recovery of edges and wide pixelwise features remains effective.*

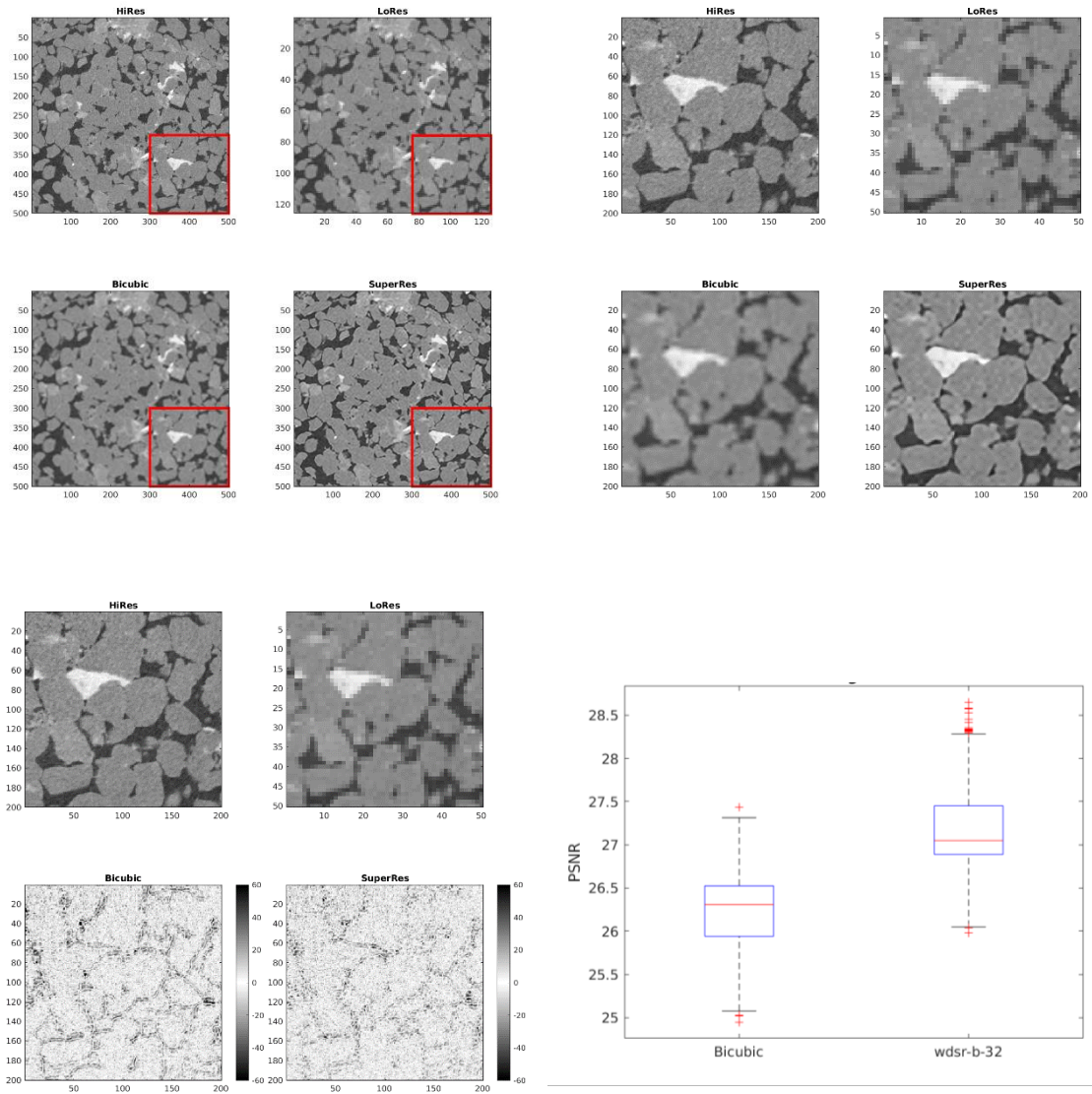

*Figure 12: SR metrics and visualisations of externally sourced Berea sandstone. The features of Berea sandstone are not present in the DRSRD1 dataset, so the model performance is impacted, seen in the bottom right boxplot. In particular, the presence of a high intensity water phase (the white features) is completely unseen by the DRSRD1 trained models.*

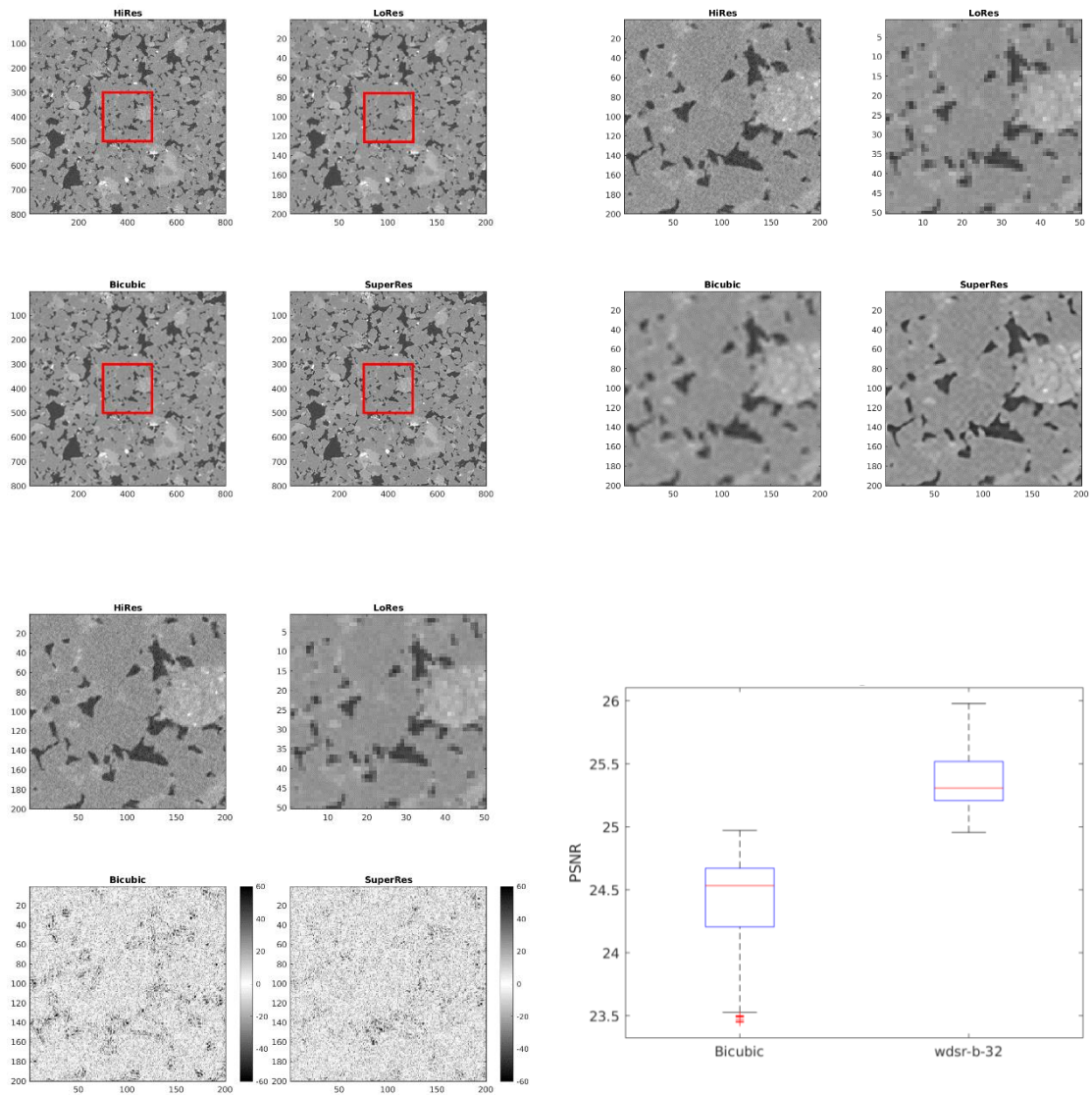

*Figure 13: SR metrics and visualisations of Herring Leopard. Model performance shows a PSNR boost of around 1 to 1.5. Visually, this can be attributed to the higher level of noise and the reduction in distinguishable features of the Leopard sandstone image due to the tighter grain packing compared to Bentheimer rock. The difference maps show that losses are still primarily noise related, so from a functional perspective the model performs well.*

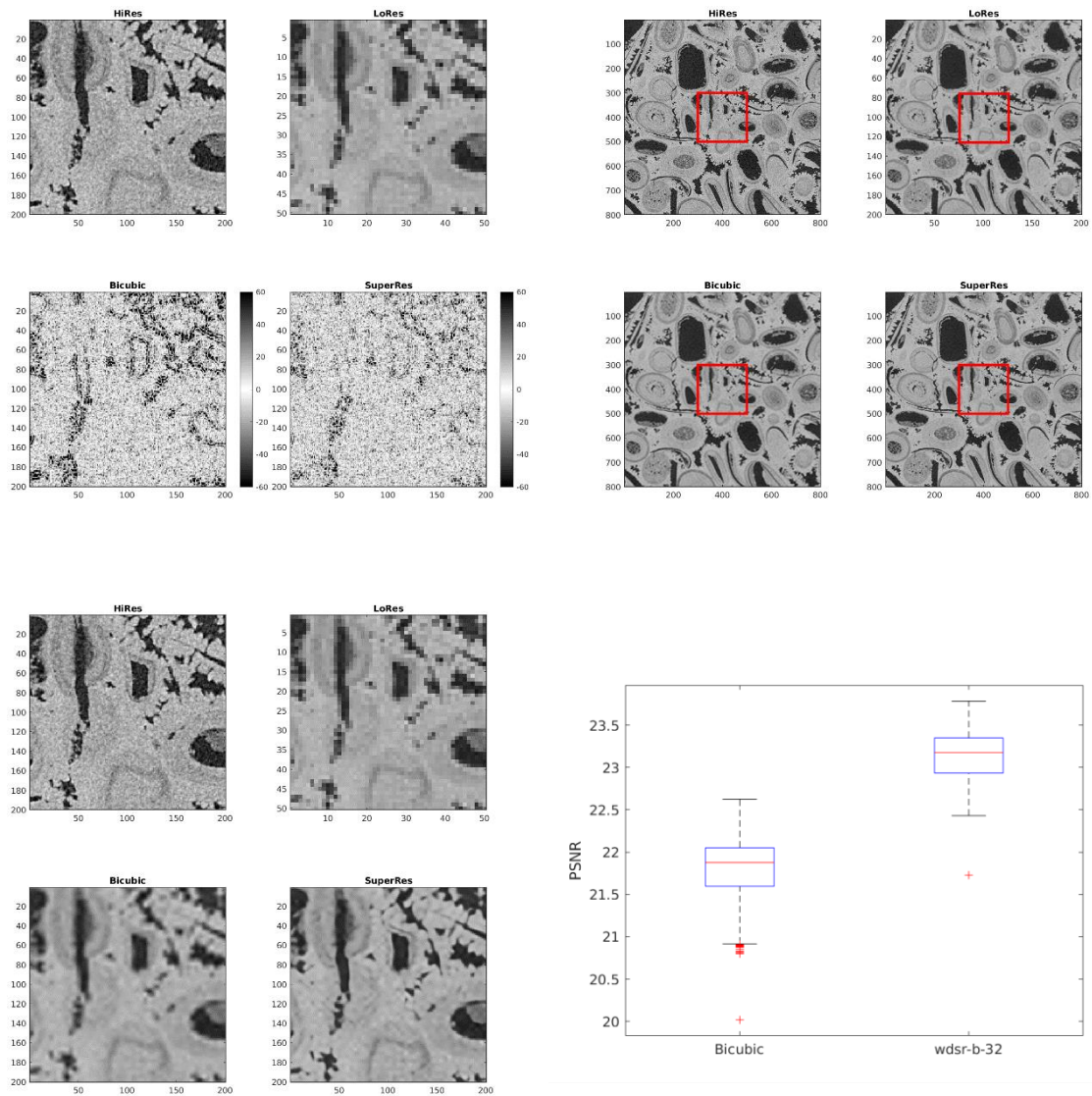

*Figure 14: SR metrics and visualisations of Savonnieres carbonate. Like the Estaillades carbonate rock in the DRSRD1 dataset, the PSNR values are low due to the increased presence of image noise. This noise contains both micro pore information and imaging noise. The lack of distinction between noise sources is not resolvable by SR methods unless the mapping itself is resolved.*

### 4.2 SUPER RESOLUTION AND AUGMENTATION OF LOW RESOLUTION micro-CT IMAGES

Despite best practices regarding generalisation of datasets to reduce overfitting and improve model flexibility, most SR benchmark datasets including DRSRD1 and DIV2K (Agustsson and Timofte 2017) (which inspired DRSDR1) are synthetic and do not account for noise and blur which are present in real LR images but absent from synthetically downsampled images. It can be expected that the addition of image noise in the LR images will impact the model training and predictive performance. In this section, the model trained on the DRSRD1 dataset (completely synthetic) is applied to generate SR images of a lower resolution (7 micron) Bentheimer rock with typical micro-CT noise.

Application of the model trained on the DRSRD1 dataset on the LR Bentheimer gives SR results (1.75 micrometres) that retain significant image noise and result in segmentation difficulty, shown in

Figure 15. On a pixel by pixel basis, this is a more accurate reconstruction as it generates high resolution noise from the low-resolution noise. This generated high-resolution noise is unwanted in DRP workflows as it impedes image segmentation. As such, it is beneficial to remove intra-granular noise during the SR process. Re-training the model on the sandstone2D dataset augmented with noise and blur (shown in Figure 17) instead results in reduced image noise while maintaining resolvable image features, seen in Figure 15. The blur was applied as a Gaussian smoothing kernel with a standard deviation randomly sampled between 0 and 1, and the noise was added to the images as a Gaussian white noise with a mean of 0 and a variance of between 0 and 0.005.

The overall pixelwise match obtained from using model trained on DRSRD1-Augmented is lower compared to using the DRSRD1 model due to the addition of essentially 'unlearnable' (or hard to learn) random noise, but visually this inadvertently produces results that show the recovery of "important features" such as grain edges, while removing intra-phase noise. This denoising is especially apparent in the histograms of the image data, shown in Figure 16, with the augmented super resolution dataset showing a significant improvement in phase contrast, potentially aiding image segmentation since visually this increase in phase contrast is consistent with the original solid-pore edge boundaries. For a digital rocks workflow, training a model to account for low resolution noise returns results that possess a natural and powerful noise filter that removes intra-phase noise and retains edge sharpness in the SR is an unforeseen benefit. This leads to a higher resolution, denoised image set for image segmentation or other greyscale workflows.

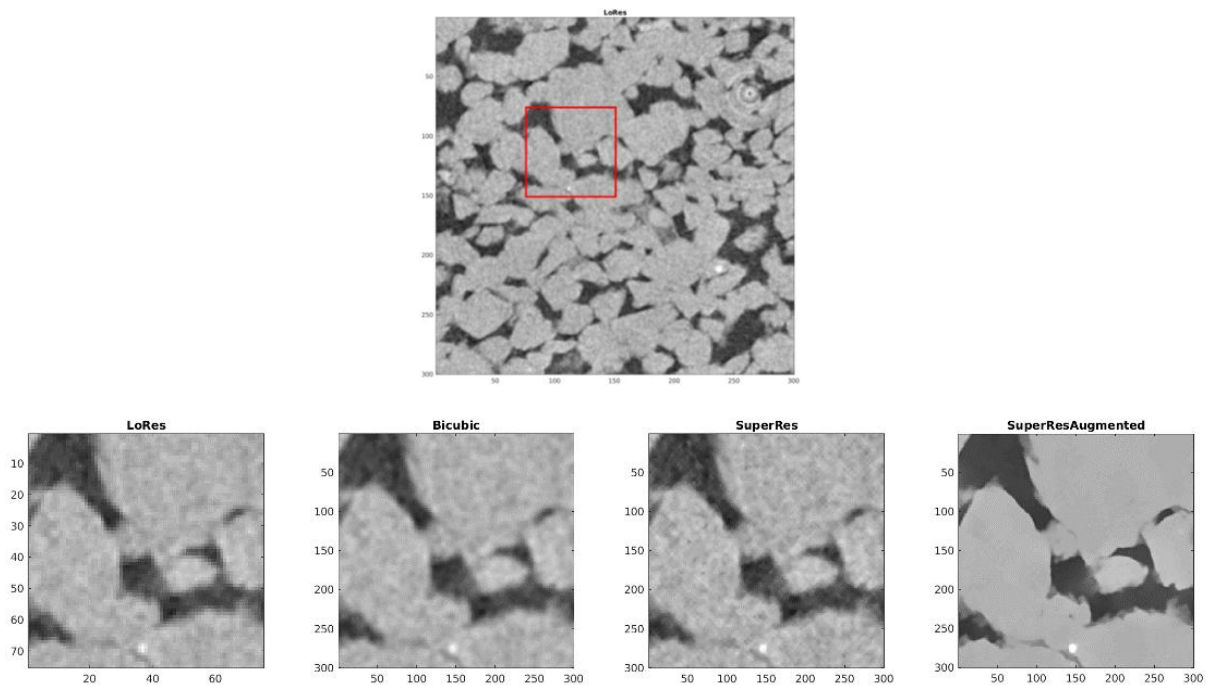

*Figure 15: Resulting SR images obtained from bicubic interpolation, the DRSRD1 trained model, and the DRSRD1 augmented trained model. The generated SR image by the DRSRD1 model (3$^{rd}$ from left) shows shaper details than the bicubic result, but the recovery of image noise from an especially noisy LR image source is problematic for digital rock workflows. The SR image generated by the DRSRD1 augmented model (right) gives a nearly completely denoised image, with all edge sharpness retained.*

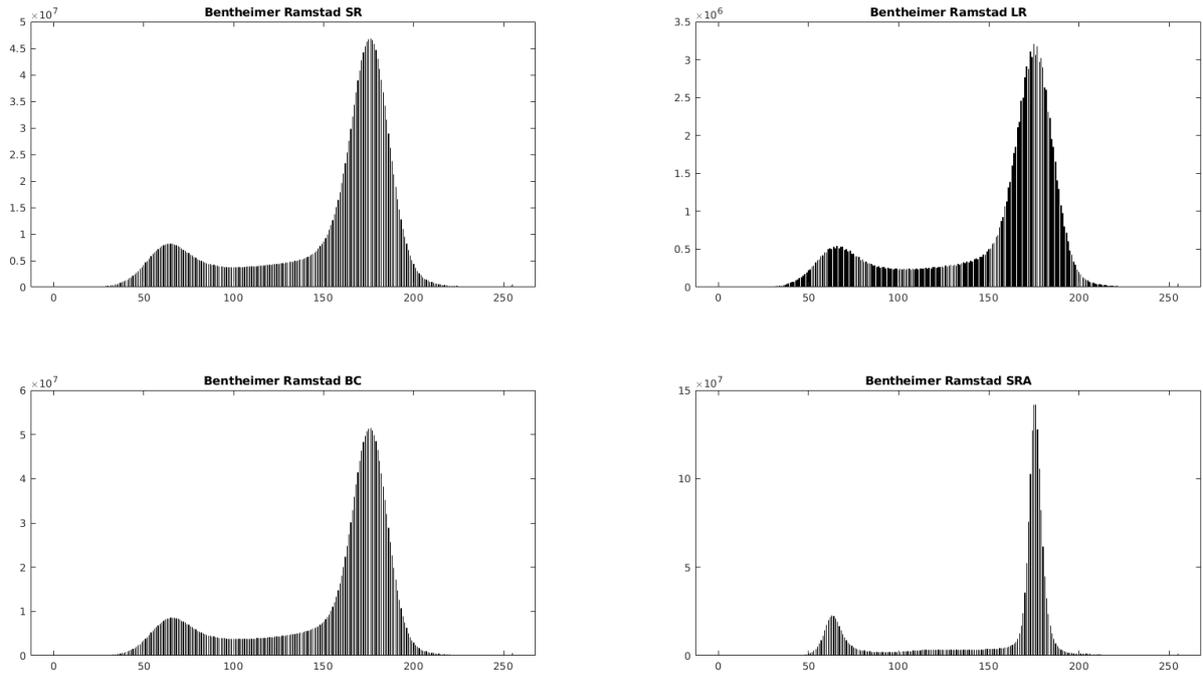

*Figure 16: Histograms of the low resolution Bentheimer sandstone, showing the augmented super resolution results with significantly better phase contrast.*

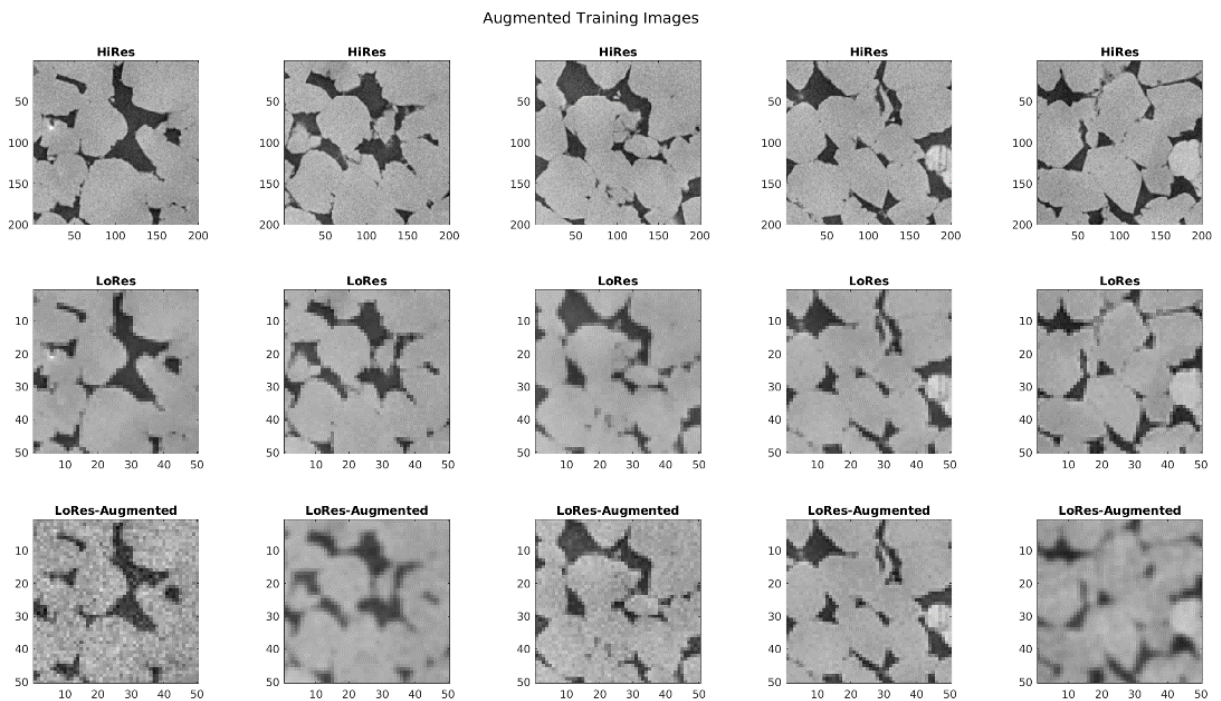

*Figure 17: Example images of augmented LR (middle row) to HR (top row) mappings with noise and blur added (bottom row)*

## 4.3 Application of Super Resolution on High Resolution Training Images

Using the augmented model that can account for real micro-CT image noise, the original sandstone HR data as well as the equivalent SR result in the DRSRD1 dataset (800x800x800) is super resolved. The high resolution image and the equivalent SR image generated for it shown in Figure 18, both measuring 800x800 at 3.8 micrometres, are super resolved to a size of 3200x3200 at 0.95 micrometres using both the original model and the augmented model used in section 4.2. This aims to investigate the generation of high-resolution images by applying SRCNN on the HR source that it is trained on as well as by applying SRCNN twice on a LR source. The zoomed in results are shown in Figure 19, and show that the obtained images share similar features as those generated in section 4.2. The HR-SR image shows that if the input image contains noise, this noise will be super-resolved and retained, resulting in an equally noisy HR-SR image, which is higher resolution, but the image quality is not improved. The SR-SR image shows that there is minimal visual improvement in the image, likely due to the increased pixelwise loss of information when applying SRCNN twice. The augmented HR-SRA and SR-SRA results on the other hand show that the edge recovery and noise suppression give superior results for the resolvable phases. The SR-SRA image compared to the HR-SRA image shows a loss in edge detail, with cusps disappearing and edges becoming smoothed. Furthermore, the noise reduction in the SR-SRA image is mottled due to the SRCNN picking up the lower frequency intra-phase fluctuations in the SR image as features of the image as opposed to high frequency noise.

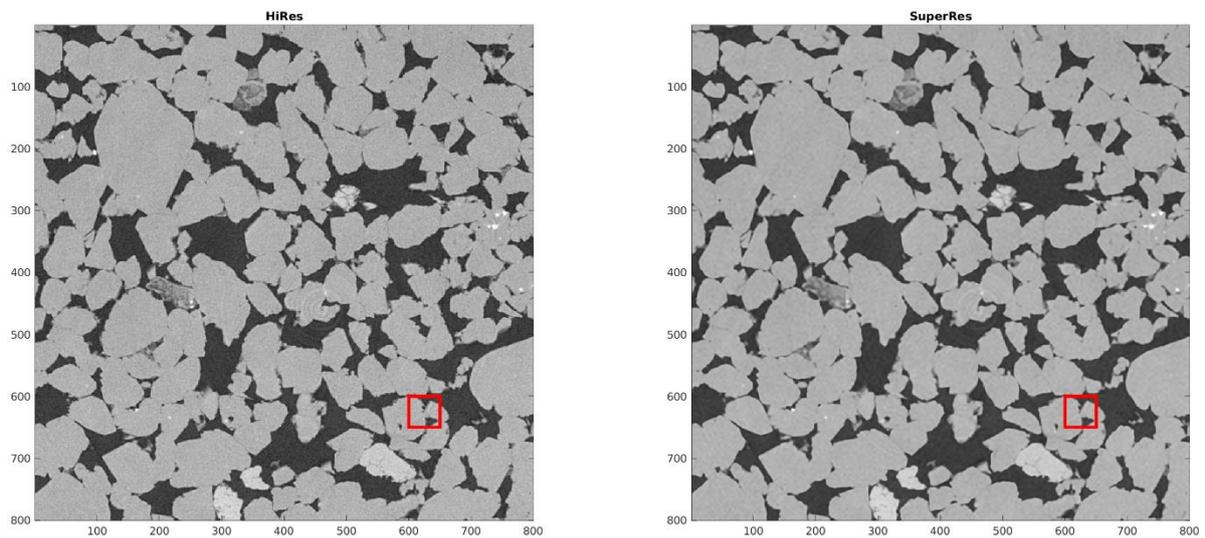

*Figure 18: HR and SR images of the Bentheimer sandstone in the DRSRD1 dataset. These images are used as an input to a further level of super resolution, resulting in an HR-SR and a SR-SR image.*

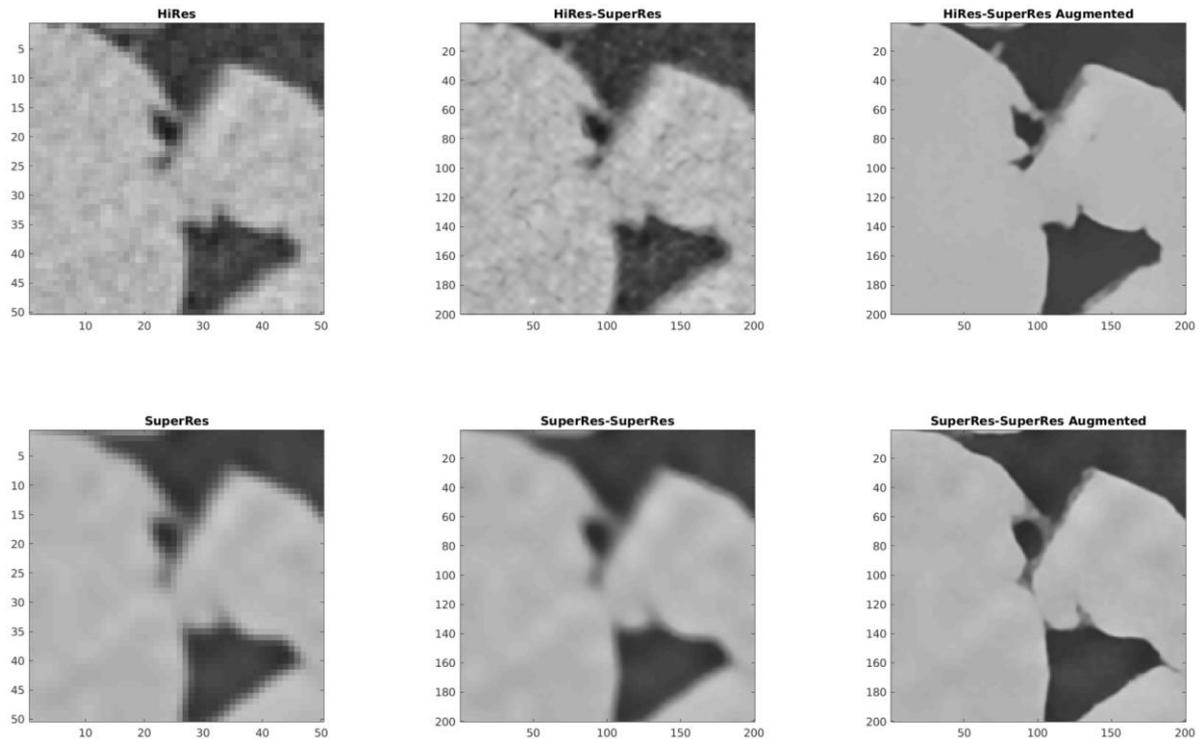

*Figure 19: SR (0.95 micrometres) images obtained by applying SRCNN on the HR source image (3.8 micrometres) and applying SRCNN twice on a LR image (15.2 micrometres). Results show a good visual consistency with previous results in section 4.2, with noise generation when using the original mode, and noise suppression when using the augmented model. The loss of edge information is apparent when applying SRCNN twice, as the bottom SR-SR images show a loss of edge detail, resulting in overly smooth, though still sharp, interfaces.*

## 5 CONCLUSIONS

Processing micro-CT images of sandstone and carbonate rocks using SRCNN as a part of Digital rock imaging has been shown to produce high quality, high resolution images that are optimised for further segmentation and grey scale analysis. The trained model PSNR compares favourably against typical interpolation methods when applied to externally sourced Bentheimer, Berea, and leopard sandstones, and Savonnieres carbonate. Difference maps indicate that edge sharpness is completely recovered in images within the scope of the trained model, with only high frequency noise related detail loss. Besides the generation of high-resolution images, a beneficial side effect of is the removal of image noise while recovering edgewise sharpness when training on augmented images with noise and blur, shown when tested against real low-resolution images of Bentheimer. The SRCNN method generates representative high-resolution images, and preconditions them for image segmentation under some circumstances, which naturally leads to future development and training of models that segment an image directly. Image restoration by SRCNN on the rock images is of significantly higher quality than traditional methods and suggests SRCNN methods are a viable processing step in a digital rock workflow.

All training was done on synthetically generated images, and though augmentation of the synthetic low-resolution images produced better super resolution outputs, a proper dataset with registered low and high resolution micro-CT data may result in better training. With the rapid development of SRCNN models, the models used in this study serve as indications of performance, as incremental improvements in the design of SRCNN layers in the future may further improve upon the results presented and discussed in this paper The use of 2D super resolution models on 3D data ignores the depth dimension, which limits this study as a proof of concept for the recovery and generation of micro-CT images that can be segmented for digital rock workflows. Application of specialised networks to 3D super resolution of micro-CT images and coupled Super Resolution with Segmentation is natural progression from this work.